\documentclass[conference]{IEEEtran}
\IEEEoverridecommandlockouts
\usepackage{cite}
\usepackage{amsmath,amssymb,amsfonts}
\usepackage{graphicx}
\usepackage{textcomp}
\usepackage{xcolor}
\usepackage{algorithm}
\usepackage{algpseudocode}
\def\BibTeX{{\rm B\kern-.05em{\sc i\kern-.025em b}\kern-.08em
    T\kern-.1667em\lower.7ex\hbox{E}\kern-.125emX}}
\begin{document}

\title{Curriculum Learning with GNN-based Reinforcement Learning for Job Shop Scheduling
{\footnotesize \textsuperscript{}}
\thanks{Supported by the Chips Joint Undertaking and its members, including top-up funding by National Authorities, within the Cynergy4MIE project (Grant Agreement No. 101140226).}
}

\author{
\IEEEauthorblockN{1\textsuperscript{st} Jayakrishnan K. Vasudevan}
\IEEEauthorblockA{\textit{Dept. of Industrial Engineering} \\
\textit{Rosenheim University} \\
\textit{of Applied Sciences} \\
Rosenheim, Germany \\
jayakrishnan.k.vasudevan@gmail.com}
\and
\IEEEauthorblockN{2\textsuperscript{nd} Jonathan Hoss}
\IEEEauthorblockA{\textit{Dept. of Industrial Engineering} \\
\textit{Rosenheim University} \\
\textit{of Applied Sciences} \\
Rosenheim, Germany \\
jonathan.hoss@th-rosenheim.de}
\and
\IEEEauthorblockN{3\textsuperscript{rd} Noah Klarmann}
\IEEEauthorblockA{\textit{Dept. of Industrial Engineering} \\
\textit{Rosenheim University} \\
\textit{of Applied Sciences} \\
Rosenheim, Germany \\
noah.klarmann@th-rosenheim.de}
}

\maketitle

\begin{abstract}
The job shop scheduling problem is a challenging combinatorial optimization problem, and recent reinforcement learning approaches using graph neural networks have shown promise for learning scheduling policies directly from problem instances. 
However, training on large instances remains computationally expensive, and generalization across instance sizes remains challenging. This paper studies curriculum learning for graph neural network-based reinforcement learning in the job shop scheduling problem by comparing it with single-size training across three target sizes: $20 \times 20$, $25 \times 25$, and $30 \times 30$. In the 
curriculum setting, the policy is first trained on smaller instances and then progressively adapted to larger target sizes, allowing scheduling behavior learned in earlier stages to support learning on larger instances. Models are evaluated on unseen instances 
from $8 \times 8$ to $30 \times 30$ using the optimality gap, considering both generalization across all evaluation sizes and specialization on the target size. Results show that curriculum learning consistently reduces wall-clock training time, with larger 
benefits as the target size increases. The strongest advantage is observed at $30 \times 30$, where curriculum learning reduces the mean optimality gap across all evaluation sizes by approximately 8.1 percentage points, reduces the target-size mean optimality gap by 
approximately 8.6 percentage points, and saves approximately 50 hours of training time.
\end{abstract}


\section{Introduction}

The Job Shop Scheduling Problem (JSSP) is a classical combinatorial optimization problem (COP) with applications in scheduling and resource allocation~\cite{smit2025graph}. Its difficulty arises from the rapidly growing number of feasible 
schedules as the number of jobs and machines increases. Since exact methods often become impractical for larger instances, a wide range of heuristic, metaheuristic, and learning-based approaches have been 
studied~\cite{gao2019fjsp_review,van2021discovering}. Recently, reinforcement learning (RL) has gained attention as a way to learn scheduling policies directly through interaction with the environment, 
reducing the dependence on manually designed dispatching rules~\cite{tassel2021reinforcement,zhang2020learning}.

Graph neural network (GNN)-based RL approaches are particularly suitable for JSSP because scheduling instances can be represented as graphs, allowing the model to capture operation precedence, machine-resource dependencies, and broader scheduling 
structure. Such models have shown promise for learning dispatching policies that generalize across instance sizes. However, direct training on large JSSP instances remains computationally expensive because larger instances lead to longer scheduling 
episodes, larger graph representations, and more complex decision dependencies~\cite{hoss2026scalable,zhang2020learning,park2021learning}. This makes training efficiency an important challenge for applying GNN-based RL to larger scheduling problems.

Curriculum learning (CL)~\cite{bengio2009curriculum,wang2021survey} offers a potential strategy for addressing this challenge by organizing training from simpler to progressively harder tasks. In JSSP, instance size is a natural curriculum variable: 
smaller instances have shorter episodes, smaller graph representations, and lower per-step computational cost, making them suitable for early training stages. A policy can therefore first learn useful scheduling behavior on smaller instances before 
being exposed to larger and more expensive target instances.

Despite growing interest in CL for COP and scheduling, several questions remain insufficiently explored. Existing JSSP curriculum studies either operate within a single problem size by ordering instances according to difficulty~\cite{de2023curriculum}, 
or mainly focus on final solution quality as the primary evaluation criterion~\cite{iklassov2022learning}. This leaves open how CL affects \textit{generalization} across unseen instance sizes, \textit{specialization} on the target size, and convergence 
behavior, including how quickly a policy improves and whether stage transitions introduce instability. In addition, wall-clock training time is rarely reported as a primary metric, although training cost is a major bottleneck for GNN-based RL on large 
JSSP instances.

This paper presents a controlled study of size-based CL for GNN-based RL in JSSP while focusing on the effect of progressively increasing instance size during training, with the aim of reducing computational cost without compromising solution quality. 
Size-based CL is compared against direct single-size training across different target sizes, and each strategy is evaluated along three axes: generalization across unseen instance sizes, specialization on the target size, and wall-clock training 
time. The central hypothesis is that, as the target instance size increases, the computational cost of direct training grows, making CL increasingly advantageous by shifting part of the learning process to earlier and computationally cheaper stages. 
Therefore, the main contributions are threefold: (1) a controlled study of single-size training versus size-based CL under identical model, evaluation, and hardware settings; (2) a separation of the evaluation into generalization across unseen instance 
sizes and specialization on the final target size, revealing trade-offs that would be hidden by a single aggregate metric; and (3) a joint analysis of wall-clock training-time savings and optimality-gap performance, showing how the advantage of CL changes 
as the target JSSP size increases.

\section{Related Work}
\label{sec:related_works}

\subsection{GNN-based RL for Scheduling}

Zhang et al.~\cite{zhang2020learning} formulated JSSP as a Markov Decision Process (MDP) over a disjunctive graph and showed that graph-based representations can support generalization across instance sizes, while Park et al.~\cite{park2021learning} 
combined RL with graph-based state encoding to capture global dependencies between operations. GNNs are well suited for JSSP because operations, precedence constraints, and machine-resource dependencies naturally form a graph structure. Through 
message passing, GNNs can learn representations that capture both local and global scheduling information~\cite{smit2025graph,battaglia2018relational}. However, graph-based RL methods for JSSP become more computationally demanding as instance size 
increases, since larger instances lead to larger graph representations and more node-embeddings \cite{park2021learning}. Therefore, scalability and wall-clock training cost remain important challenges for GNN-based RL scheduling methods.

\subsection{CL for COP}

CL, as formalized by Bengio et al.~\cite{bengio2009curriculum}, is built on the principle that models learn more effectively when training examples are ordered from simple to complex, rather than presented uniformly at random. This idea has since 
been extended to RL, where curricula organize the agent's experience as a structured progression of tasks or task distributions, improving both learning stability and efficiency~\cite{narvekar2020curriculum,klink2022curriculum}.

Several studies have explored curriculum strategies for COP. Waubert de Puiseau et al.~\cite{de2023curriculum} order training by \textit{instance difficulty} for JSSP instances within a fixed problem size, relying on hand-crafted difficulty measures 
based on heuristics-solutions to rank individual examples from easy to hard and vice versa. Lisicki et al.~\cite{lisicki2020evaluating} investigated \textit{problem size} as the curriculum dimension by progressively exposing an attention-based model 
to larger Traveling Salesman Problem instances. They introduced an adaptive staircase curriculum, in which the difficulty level is updated during training by allowing the model to move to an easier level, stay at the current level, or advance to a 
more difficult one based on its performance. Iklassov et al.~\cite{iklassov2022learning} further explore this curriculum design by reinforcing the curriculum  with problems where the model performs poorly, thereby focusing training on more challenging 
cases. These two strategies in ~\cite{de2023curriculum} and ~\cite{iklassov2022learning, lisicki2020evaluating} reflect different assumptions about what constitutes a useful learning progression.

\section{Preliminaries}
\label{sec:preliminaries}

\subsection{JSSP and RL Formulation}

The JSSP consists of jobs $J=\{J_1,\ldots,J_n\}$ and machines $M=\{M_1,\ldots,M_m\}$. Each job $J_j$ is composed of an ordered set of operations $(O_{j,1},\ldots,O_{j,k_j})$, where operation $O_{j,k}$ must be processed on a specified machine for processing 
time $p_{j,k}$. A schedule is feasible if every operation is assigned a start time $t_{j,k}^{\mathrm{start}}$ while respecting both job precedence, $t_{j,k+1}^{\mathrm{start}} \geq t_{j,k}^{\mathrm{start}}+p_{j,k}$, and machine-capacity constraints, such that 
no machine processes more than one operation simultaneously. The goal is to minimize the makespan $C_{\max}=\max_{j\in J}(t_{j,k_j}^{\mathrm{start}}+p_{j,k_j})$.

In this study, the JSSP is modeled as an MDP $(\mathcal{S},A,P,R,\gamma)$, where $\mathcal{S}$ denotes the state space, $A$ the action space, $P$ the transition dynamics, $R$ the reward function, and $\gamma$ the discount factor ~\cite{sutton1998reinforcement}. 
The agent constructs a schedule sequentially by selecting one eligible operation at each decision step. The state $s_t \in \mathcal{S}$ represents the current scheduling status, including completed operations, available operations, machine availability, and 
remaining work. The action $a_t \in A(s_t)$ is selected from the feasible action set obtained through action masking. After an action is selected, the next state is determined by $P$ according to precedence and machine availability constraints. 
The policy $\pi_\theta(a_t|s_t)$ learns which operation to select next.

To evaluate solution quality, this study uses the optimality gap (OG) with respect to the Flow Due Date/Most Work Remaining (FDD/MWR) heuristic, which serves as the reference method for comparison and it is defined as

\[
OG = \frac{C_{\text{model}} - C_{\text{FDD/MWR}}}{C_{\text{FDD/MWR}}},
\]

where $C_{\text{model}}$ is the makespan produced by the learned policy and $C_{\text{FDD/MWR}}$ is the makespan obtained using the FDD/MWR heuristic. This metric quantifies the relative difference between the learned policy and the reference heuristic. A 
lower OG indicates better solution quality, with values close to zero indicating performance similar to FDD/MWR.

\section{Methodology}
\label{sec:methodology}

\subsection{GNN-Based RL Framework and Graph Representation}

This study adopts the linear-complexity graph-based RL scheduling architecture proposed by Hoss et al. \cite{hoss2026scalable} as the policy model $\pi_\theta$ for the experiments. The model follows a constructive scheduling formulation, in which the 
schedule is built sequentially by selecting one operation at each decision step. The current shop-floor state is encoded as a graph, processed by a GNN backbone, and used by an actor head to select the next feasible operation. The actor--critic policy is 
trained using Proximal Policy Optimization (PPO) \cite{schulman2017proximal}, with $\theta$ denoting the actor parameters and $\phi$ denoting the critic parameters.

At each decision step $t$, the scheduling state is represented as an operation--machine graph $G_t=(V,E)$, where $V = V_{\mathrm{ops}} \cup V_{\mathrm{mch}}$ consists of operation nodes and machine nodes. The edge 
set $E = E_{\mathrm{prec}} \cup E_{\mathrm{assign}}$ contains directed precedence edges between consecutive operations of the same job and bidirectional assignment edges between operations and their required machines. 
Although the graph is structurally heterogeneous, feature-based homogenization maps operations and machine nodes into a shared feature space using node-type indicators and dynamic scheduling features. This allows the graph to be processed by a 
homogeneous Graph Isomorphism Network (GIN) \cite{xu2018powerful} while preserving the distinction between operations and machines. The action masking ensures that only precedence-feasible operations are selected. The actor selects the next operation 
using action masking, while the critic estimates the value of the current scheduling state. For all experiments, a fixed GNN architecture is used based on the configuration adopted from~\cite{hoss2026scalable}; the corresponding hyperparameters are 
summarized in Table~\ref{tab:hyperparameter_configuration}.

\begin{table}[t]
\centering
\caption{GNN Backbone and Policy Head Hyperparameters}
\label{tab:hyperparameter_configuration}
\begin{tabular}{lc}
\hline
\textbf{Parameter} & \textbf{Value} \\
\hline
\multicolumn{2}{l}{\textit{Architecture}} \\
Hidden dimension & 64 \\
Number of GIN layers ($K$) & 3 \\
Input feature dimension & 5 \\
\hline
\multicolumn{2}{l}{\textit{Training}} \\
Learning rate & $3 \times 10^{-4}$ with linear decay\\
Batch size & $5 \times (|J| \times |M|)$ transitions \\
Discount factor ($\gamma$) & 0.995 \\
GAE parameter ($\lambda$) & 1.0 \\
Clip coefficient ($\epsilon$) & 0.2 \\
Entropy coefficient ($c_{\mathrm{ent}}$) & 0.001 \\
\hline
\end{tabular}
\end{table}

\subsection{CL Training Strategies}
\label{sec:cl_training_strategies}

This work investigates CL as a training strategy for GNN-based RL in JSSP. Instead of training directly on the final target size, CL exposes the policy to a sequence of progressively larger instance sizes.  In this study, the curriculum consists of three 
stages with instance sizes $\mathcal{C}=[c_1,c_2,c_3]$, where $c_1 < c_2 < c_3$ and $c_3$ is the target size. Here, $c_k$ denotes the JSSP instance size used at curriculum stage $k$. Each stage $k$ is allocated a fixed budget of $T_{\mathrm{stage}}=10$ 
million (M) environment steps, giving a total training budget of $T_{\max}=3\cdot T_{\mathrm{stage}}=30\mathrm{M}$ steps, equal to the single-size baseline.

At curriculum stage $k$, the environment $\mathcal{E}$ is configured for instance size $c_k$. The agent interacts with $\mathcal{E}$ by executing policy $\pi_\theta$ to collect a batch of trajectories $\mathcal{D}=\{\tau_i\}$ until the rollout buffer is filled. The policy parameters $\theta$ and value-function parameters $\phi$ are
then updated via PPO. This collect--update cycle repeats until the per-stage step budget $k\cdot T_{\mathrm{stage}}$ is exhausted. At every $\Delta_{\mathrm{eval}}$ steps, $\pi_\theta$ is evaluated on the fixed held-out evaluation dataset.

When transitioning from stage $k$ to $k+1$, the environment $\mathcal{E}$, rollout collector, and all size-dependent hyperparameters are reconfigured for $c_{k+1}$ while the parameters $(\theta,\phi)$ are \emph{retained}, so the scheduling behavior learned 
at $c_k$ is transferred to the next stage.

A three-stage curriculum is used because fewer stages may not provide sufficient transition from smaller to larger instances, while more stages would introduce additional design choices beyond the scope of this study. The equal allocation of 
$T_{\mathrm{stage}}=10\mathrm{M}$ steps per stage keeps the schedule symmetric across all target sizes. Overall, this design provides a simple and symmetric size progression while keeping the curriculum schedule directly comparable to the single-size 
baseline under the same total training budget. The full procedure is given in Algorithm~\ref{alg:curriculum_learning}.

\begin{algorithm}[t]
\caption{Three-Stage CL for JSSP}
\label{alg:curriculum_learning}
\begin{algorithmic}[1]
\Require Curriculum stage sizes $\mathcal{C} = [c_1, c_2, c_3]$, stage budget $T_{\mathrm{stage}}$, total budget $T_{\max} = 3 \cdot T_{\mathrm{stage}}$, evaluation interval $\Delta_{\mathrm{eval}}$
\Ensure Trained policy $\pi_\theta$
\State Initialize policy parameters $\theta$, value-function parameters $\phi$, step counter $t \leftarrow 0$
\For{$k = 1$ \textbf{to} $3$}
    \State Configure environment $\mathcal{E}$, rollout collector, and size-dependent hyperparameters for instance size $c_k$
    \While{$t < k \cdot T_{\mathrm{stage}}$}
        \State Collect trajectories $\mathcal{D} \leftarrow \{\tau_i\}$ using $\pi_\theta$ until rollout buffer is full
        \State Update $(\theta, \phi) \leftarrow \mathrm{PPO}(\mathcal{D}, \theta, \phi)$
        \State $t \leftarrow t + |\mathcal{D}|$
        \If{$t \bmod \Delta_{\mathrm{eval}} = 0$}
            \State Evaluate $\pi_\theta$ on fixed held-out evaluation dataset
        \EndIf
    \EndWhile
    \If{$k < 3$}
        \State Adapt $\mathcal{E}$ and rollout collector to $c_{k+1}$; retain $\theta$ and $\phi$  
    \EndIf
\EndFor
\end{algorithmic}
\end{algorithm}

\section{Experimental Setup}
\label{sec:exp_setup}

\subsection{Instance Sizes and Problem Setting}

Instances with equal numbers of jobs and machines ($n = m$, i.e., a job-to-machine ratio $J/M = 1$) are used throughout the study. This choice is motivated by prior scheduling studies suggesting that instances with $J/M \approx 1$ tend to exhibit more 
difficult scheduling landscapes than other configurations~\cite{streeter2006landscape}. It is also consistent with the structural saturation hypothesis proposed in our recent work~\cite{hoss2026scalable}, which suggests that scheduling difficulty is governed 
more strongly by problem topology and constraint density than by absolute instance size. In particular, the regime $J/M = 1$ represents a highly saturated setting in which machine contention is sufficiently pronounced to require robust conflict-resolution 
strategies. Moreover, restricting both training and evaluation to $J/M = 1$ provides a controlled setting for isolating the effect of increasing instance size. Therefore, observed learning behavior can be attributed mainly to problem size rather than changes 
in the job-to-machine ratio. The training instances follow the random JSSP generation protocol used in~\cite{hoss2026scalable}: for each job, the machine route is sampled as a random permutation of the machines, and the operation processing times are drawn 
independently as $p_{j,k} \sim \mathcal{U}\{1, 2, \ldots, 99\}$.

Experiments are conducted for three target sizes: $20 \times 20$, $25 \times 25$, and $30 \times 30$. The selection of $20 \times 20$ is motivated by the GNN-based RL architecture adopted in this study. In~\cite{hoss2026scalable}, the $20 \times 20$ 
configuration was identified as a structurally important training scale that supports strong zero-shot generalization across different instance sizes and shapes. Therefore, $20 \times 20$ is used as the base target size for comparing CL with direct 
single-size training.

The larger target sizes, $25 \times 25$ and $30 \times 30$, are selected to increase problem complexity. This allows the study to examine whether the benefits of CL remain consistent or become more critical as the target instance size grows. 
Since these target sizes make direct single-size training computationally expensive, they are suitable for evaluating training efficiency.

\subsection{Training Procedure}

For each target size, CL is compared against a corresponding single-size training baseline. In single-size training, the policy $\pi_\theta$ is trained exclusively on the target instance size for the full budget of $T_{\max}=30\mathrm{M}$ environment steps. 
In CL, the same total budget is divided into three stages of $T_{\mathrm{stage}}=10\mathrm{M}$ steps, where the final curriculum stage corresponds to the same target size as the single-size baseline.

Both strategies use the same model architecture, hyperparameters, and hardware configuration, consisting of an AMD Ryzen 9 7950X3D 16-core processor with 32 CPU threads. They are also evaluated at the same intervals using the same fixed evaluation dataset 
described in Section~\ref{sec:evaluation_dataset}. This controlled setup ensures that observed differences can be attributed primarily to the training strategy and target instance size.

The specific training configurations are shown in Table~\ref{tab:training_configurations}. The curriculum model ending at $20 \times 20$ is compared to a model trained exclusively on $20 \times 20$ instances. Similarly, the $25 \times 25$ and $30 \times 30$ 
curriculum runs are compared against their corresponding single-size baselines. For the target size $20 \times 20$, the curriculum proceeds through $10 \times 10$, $15 \times 15$, and finally $20 \times 20$ instances. For the target size $25 \times 25$, the 
model is first trained on $10 \times 10$ and $15 \times 15$ instances before moving to $25 \times 25$. For the target size $30 \times 30$, the curriculum uses $10 \times 10$ and $20 \times 20$ as intermediate stages before training on $30 \times 30$. Thus, 
smaller problem sizes are used as curriculum stages intended to initialize learning before training on larger target instances.

\begin{table}[t]
\centering
\caption{Training Configurations}
\label{tab:training_configurations}
\resizebox{\columnwidth}{!}{
\begin{tabular}{|c|c|c|c|}
\hline
\textbf{Target size} & \textbf{Single-size} & \textbf{CL} & \textbf{Total steps} \\
\hline
$20 \times 20$ & $20 \times 20$ & $10 \times 10 \rightarrow 15 \times 15 \rightarrow 20 \times 20$ & 30M \\
\hline
$25 \times 25$ & $25 \times 25$ & $10 \times 10 \rightarrow 15 \times 15 \rightarrow 25 \times 25$ & 30M\\
\hline
$30 \times 30$ & $30 \times 30$ & $10 \times 10 \rightarrow 20 \times 20 \rightarrow 30 \times 30$ & 30M\\
\hline
\end{tabular}
}
\end{table}

To reduce the effect of stochasticity inherent in RL training, each experimental setting is repeated over three independent runs with different random seeds. The reported curves and final values correspond to the mean performance across these runs. 
The shaded regions in the learning curves indicate one standard deviation across the three independent runs, providing an estimate of run-to-run variability and the consistency of the observed trends.

\subsection{Evaluation Dataset}
\label{sec:evaluation_dataset}

Periodic evaluation is performed every $\Delta_{\mathrm{eval}}$ steps throughout training using a fixed held-out evaluation dataset. This dataset contains 20 instances for each evaluation size, $8 \times 8$, $10 \times 10$, $12 \times 12$, $15 \times 15$, 
$20 \times 20$, $25 \times 25$, and $30 \times 30$, resulting in 140 evaluation instances in total. The evaluation instances are generated from the same distribution and using the same random JSSP generation procedure as the training instances. However, 
a different random seed is used for generating evaluation instances, ensuring that these instances are disjoint from the training instances. Evaluating at regular intervals rather than only at the end of training  enables comparison of convergence behavior 
between single-size training and CL. 

\subsection{Evaluation Metrics}

This study focuses on three primary evaluation metrics, as summarized in Table~\ref{tab:evaluation_metrics}.

\begin{table}[t]
\centering
\caption{Evaluation Metrics}
\label{tab:evaluation_metrics}
\resizebox{\columnwidth}{!}{
\begin{tabular}{|c|c|c|}
\hline
\textbf{Metric} & \textbf{Evaluated on} & \textbf{Purpose} \\
\hline
$\overline{OG}_{\mathrm{all}}$ & Sizes from $8 \times 8$ to $30 \times 30$ & Generalization \\
\hline
$\overline{OG}_{\mathrm{target}}$ & Target instance size only & Specialization \\
\hline
Wall-clock training time & Full training process & Training efficiency \\
\hline
\end{tabular}
}

\end{table}

\begin{itemize}
    \item \textit{Generalization performance:} This is measured using the average optimality gap across all evaluation sizes, defined as $\overline{OG}_{\mathrm{all}} = \frac{1}{N}\sum_{i=1}^{N} OG_i$, where $N$ is the total number of evaluation instances 
    across all sizes. This metric reflects how well the trained model performs across a broad range of unseen instance sizes.

    \item \textit{Specialization performance:} This is measured using the average optimality gap on the target training size, defined as $\overline{OG}_{\mathrm{target}} = \frac{1}{N_{\mathrm{target}}}\sum_{i=1}^{N_{\mathrm{target}}} OG_i$, where 
    $N_{\mathrm{target}}$ is the number of evaluation instances of the target size. For example, in the $20 \times 20$ experiment, this metric is computed only on the $20 \times 20$ evaluation instances.

    \item \textit{Training efficiency:} This is measured using wall-clock time, defined as the actual elapsed time required to complete training. This includes environment interaction, policy optimization, logging, and periodic evaluation. The time saved 
    by CL is computed relative to the corresponding single-size training baseline.
\end{itemize}

\section{Results and Discussion}
\label{sec:results}

This section compares CL against single-size training for the three target sizes. The comparison focuses on two optimality-gap metrics and training time as discussed in the previous section. For each target size, four types of curves are used:
\begin{itemize}
    \item Mean OG across all evaluation sizes vs. training steps.
    \item Mean OG across all evaluation sizes vs. wall-clock time.
    \item Mean OG on the target size vs. training steps.
    \item Mean OG on the target size vs. wall-clock time.
\end{itemize}

These curves allow the comparison to capture both solution quality and training efficiency.

\subsection{Results for $20 \times 20$ Target Size}

\begin{figure}[t]
    \centering

    \includegraphics[
        width=\columnwidth,
        height=0.22\textheight,
        keepaspectratio
    ]{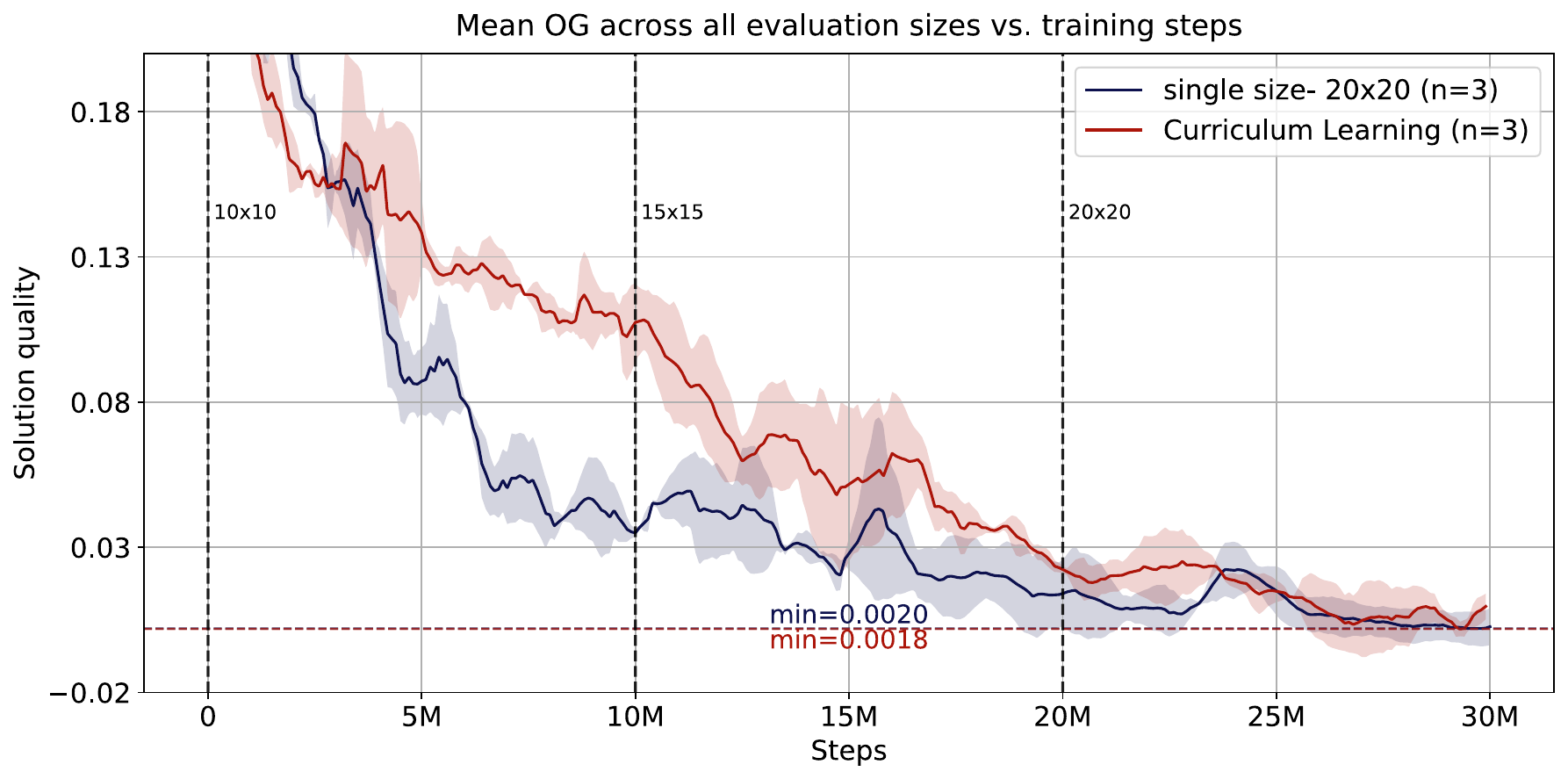}

    \vspace{1mm}

    \includegraphics[
        width=\columnwidth,
        height=0.22\textheight,
        keepaspectratio
    ]{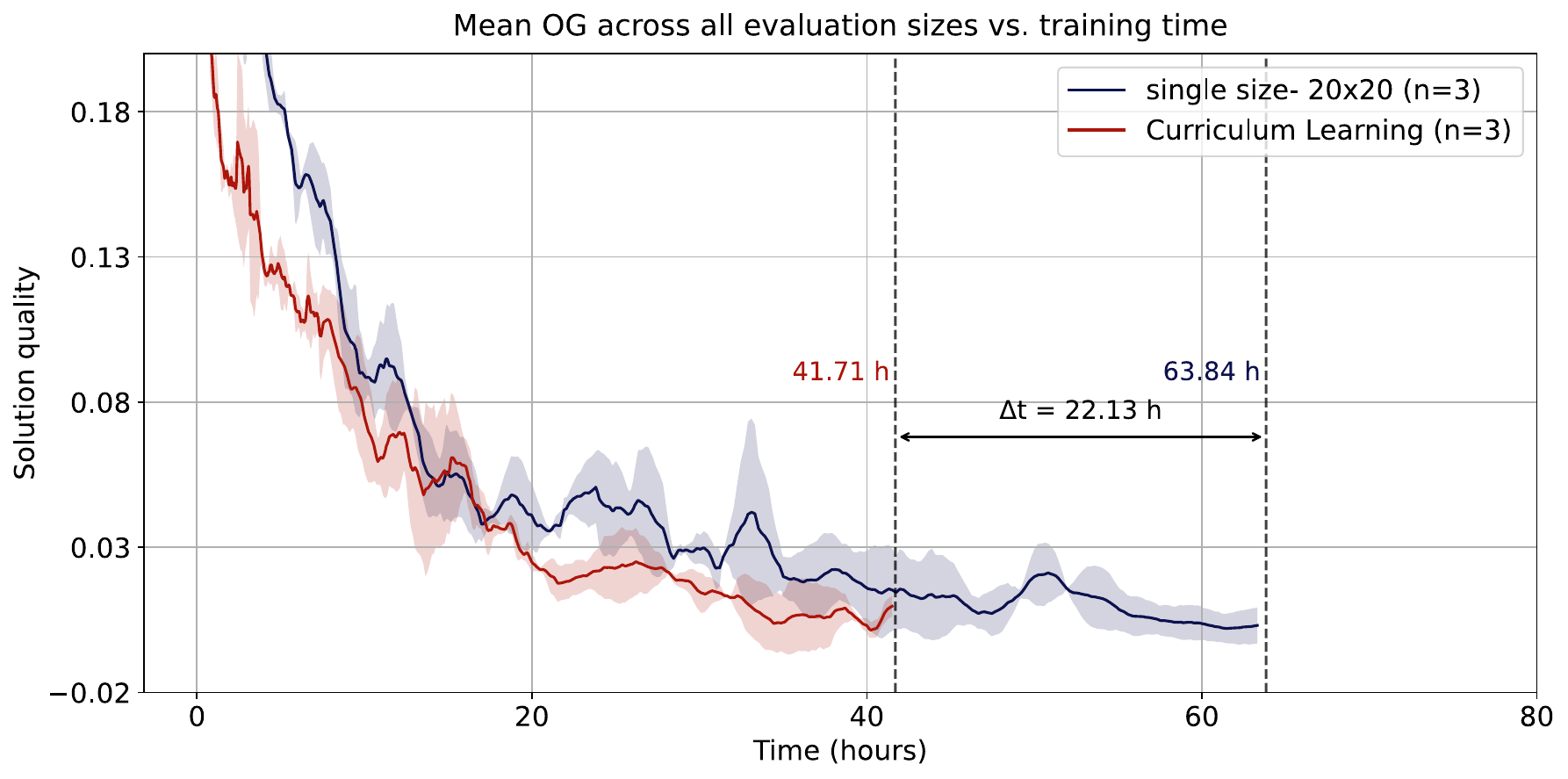}

    \caption{Generalization performance for the $20 \times 20$ target setting. Top: mean OG across all evaluation sizes versus training steps. Bottom: mean OG across all evaluation sizes versus wall-clock training time.}
    \label{fig:generalization_20}
\end{figure}

\begin{figure}[t]
    \centering

    \includegraphics[
        width=\columnwidth,
        height=0.22\textheight,
        keepaspectratio
    ]{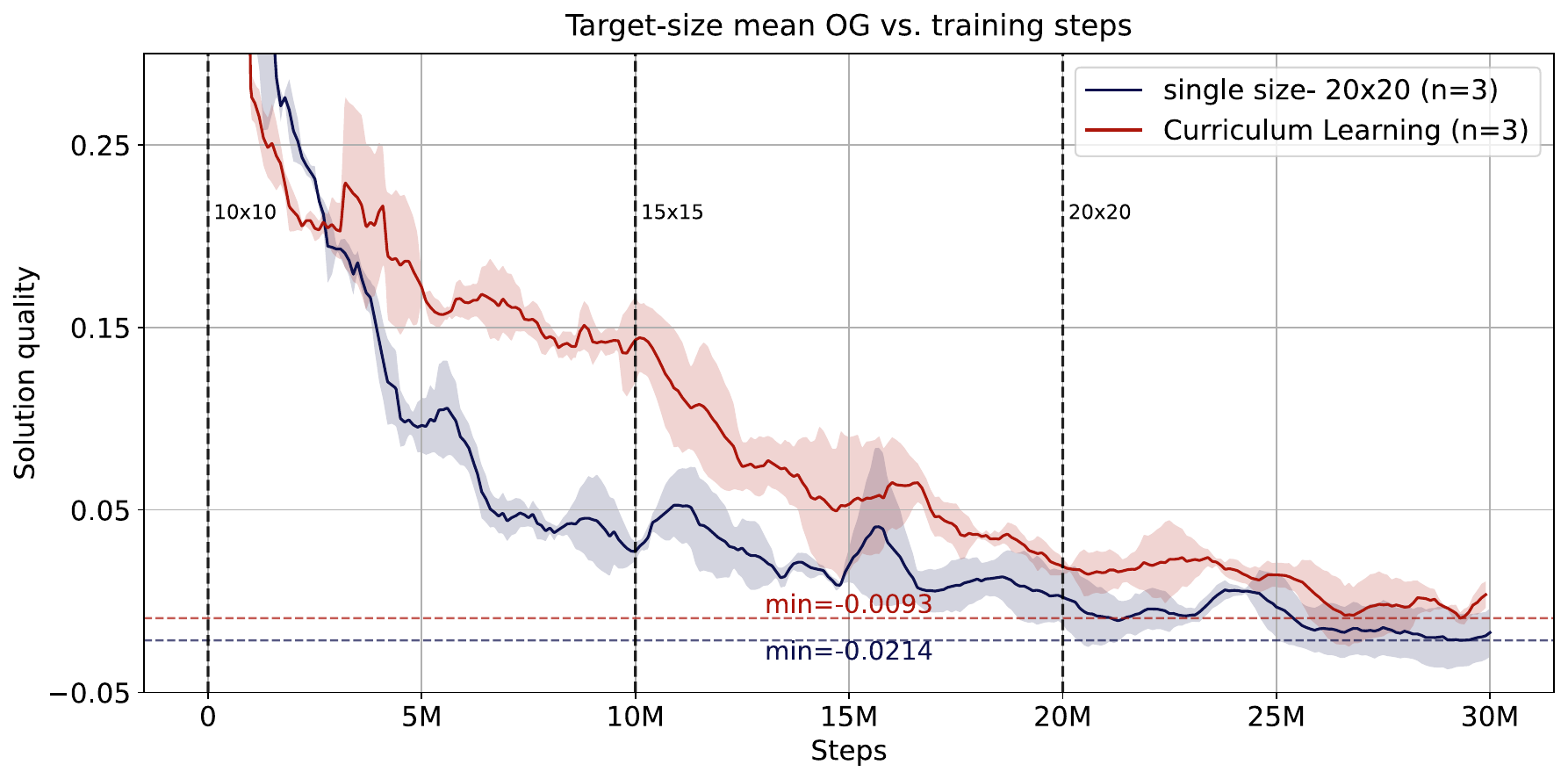}

    \vspace{1mm}

    \includegraphics[
       width=\columnwidth,
        height=0.22\textheight,
        keepaspectratio
    ]{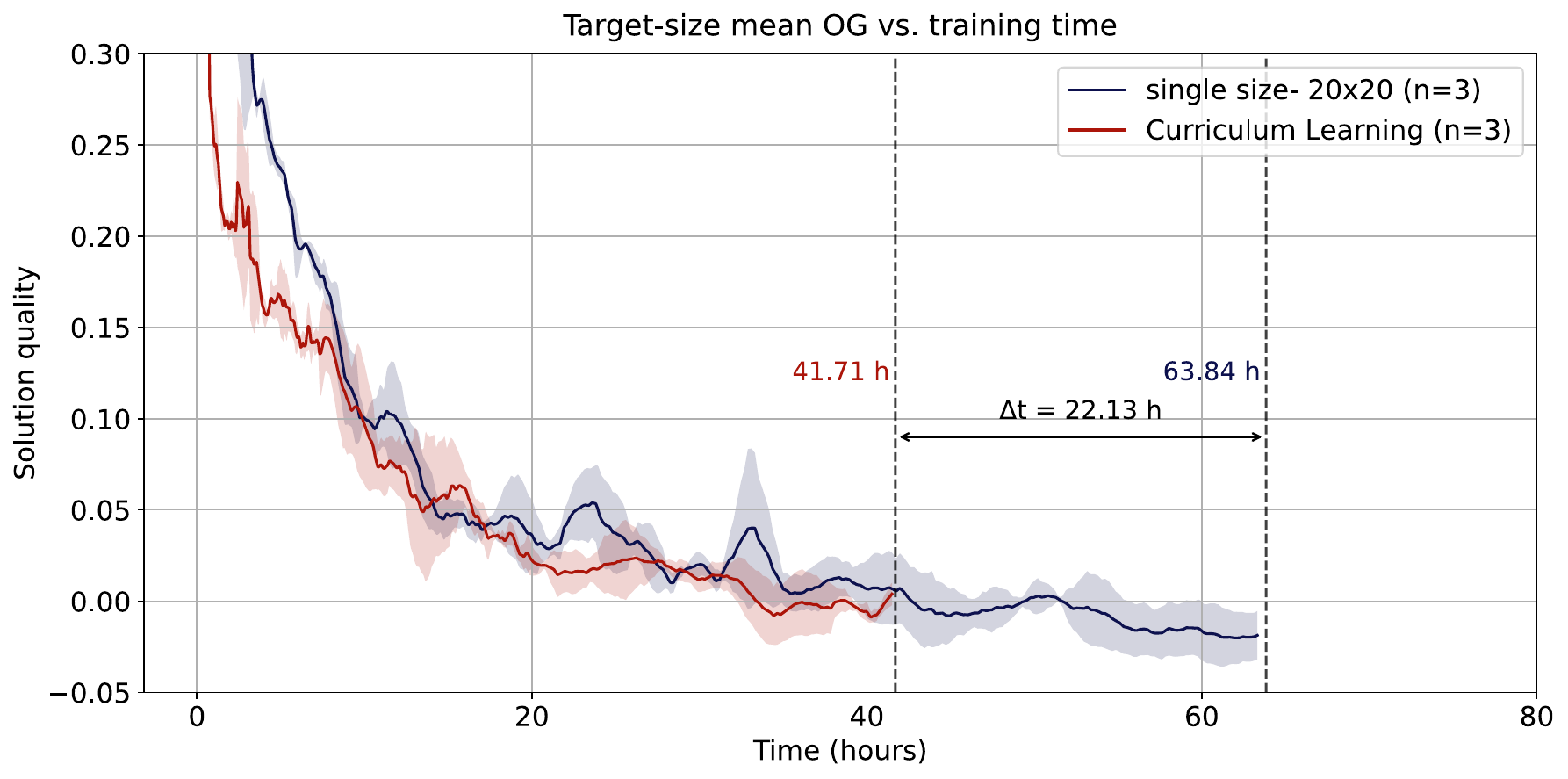}

    \caption{Specialization performance for the $20 \times 20$ target setting. Top: mean OG on $20 \times 20$ evaluation instances versus training steps. Bottom: mean OG on $20 \times 20$ evaluation instances versus wall-clock training time.}
    \label{fig:specialization_20}
\end{figure}

For the $20 \times 20$ setting, CL provides a clear training-efficiency advantage while maintaining similar generalization performance. As shown in Fig.~\ref{fig:generalization_20}, CL reduces wall-clock training time by approximately 20 hours, 
corresponding to about a 34\% reduction compared to single-size training. In terms of mean OG across all evaluation sizes, CL performs similarly to single-size training, indicating that training first on smaller instances does not substantially 
affect performance on the mixed evaluation set. When the evaluation is restricted to the target size, CL shows a small specialization loss, as shown in Fig.~\ref{fig:specialization_20}. Specifically, CL increases the target-size mean OG by approximately 
1.2 percentage points compared to the single-size model. This suggests that direct training on $20 \times 20$ instances provides a slight advantage when specializing on that specific size. Overall, CL offers a favorable efficiency benefit and comparable generalization 
performance, at the cost of a small specialization trade-off.

\subsection{Results for $25 \times 25$ Target Size}

\begin{figure}[t]
    \centering

    \includegraphics[
        width=\columnwidth,
        height=0.22\textheight,
        keepaspectratio
    ]{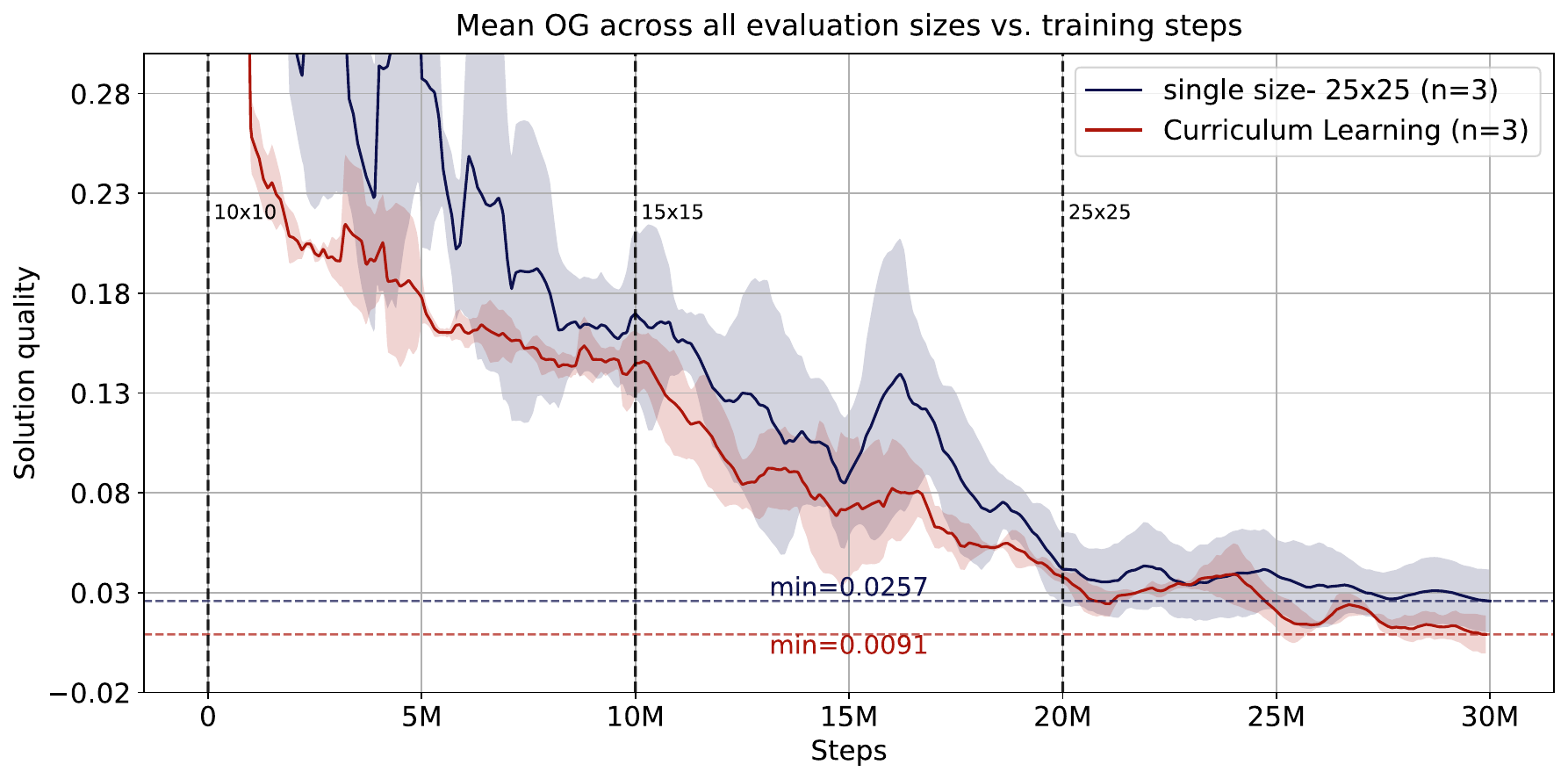}

    \vspace{1mm}

    \includegraphics[
        width=\columnwidth,
        height=0.22\textheight,
        keepaspectratio
    ]{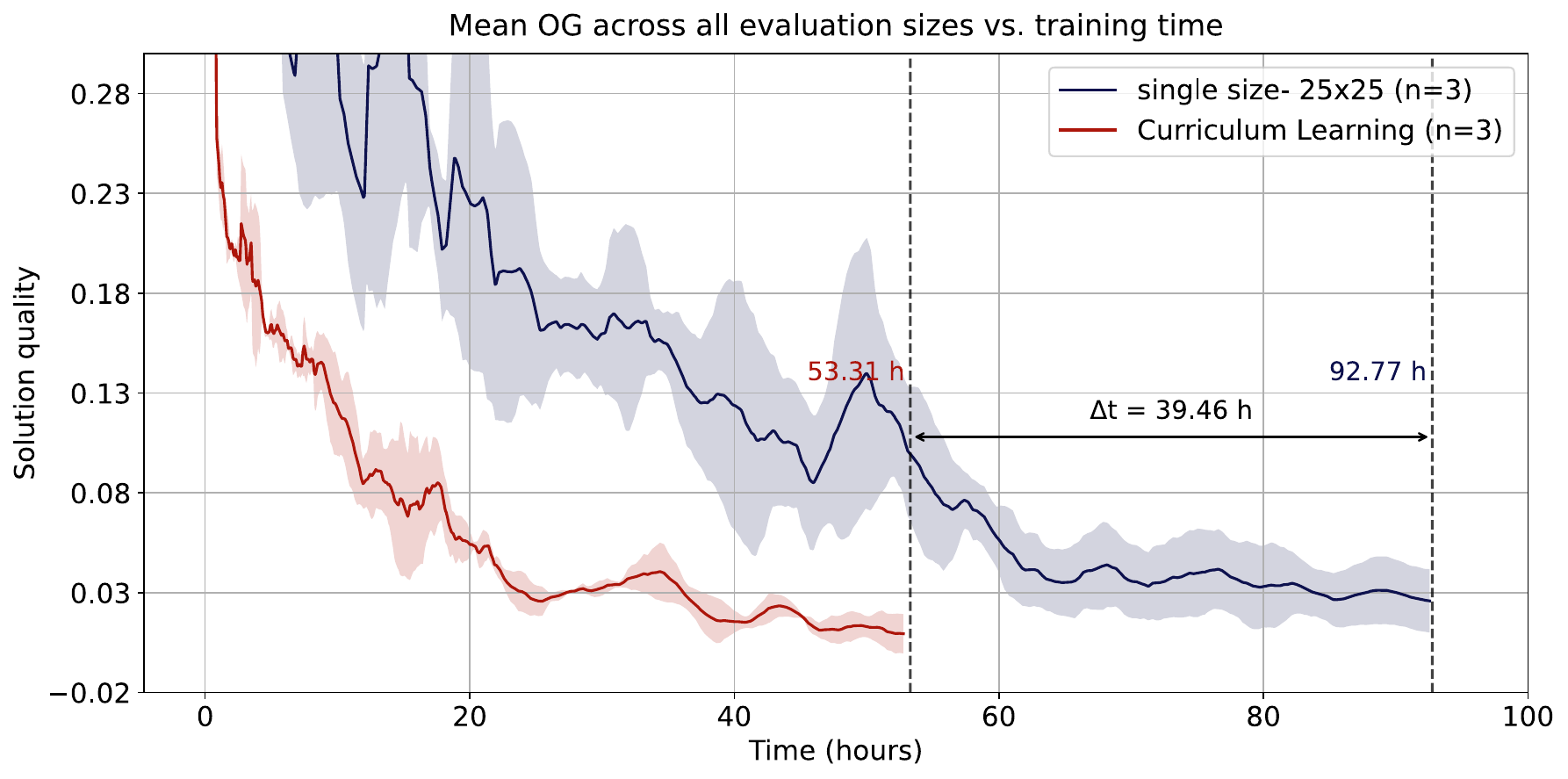}

    \caption{Generalization performance for the $25 \times 25$ target setting. Top: mean OG across all evaluation sizes versus training steps. Bottom: mean OG across all evaluation sizes versus wall-clock training time.}
    \label{fig:generalization_25}
\end{figure}

\begin{figure}[t]
    \centering

    \includegraphics[
        width=\columnwidth,
        height=0.22\textheight,
        keepaspectratio
    ]{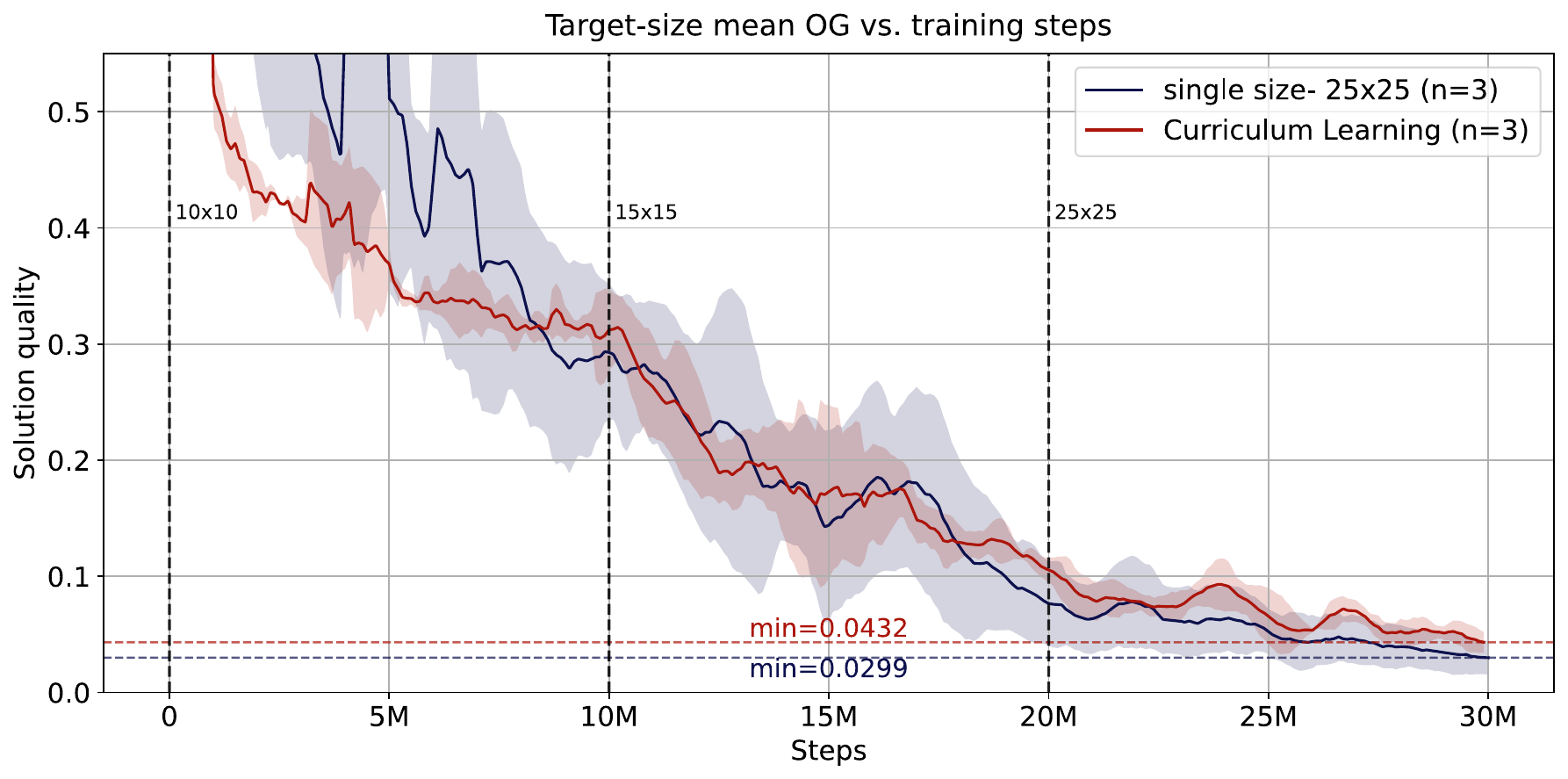}

    \vspace{1mm}

    \includegraphics[
        width=\columnwidth,
        height=0.22\textheight,
        keepaspectratio
    ]{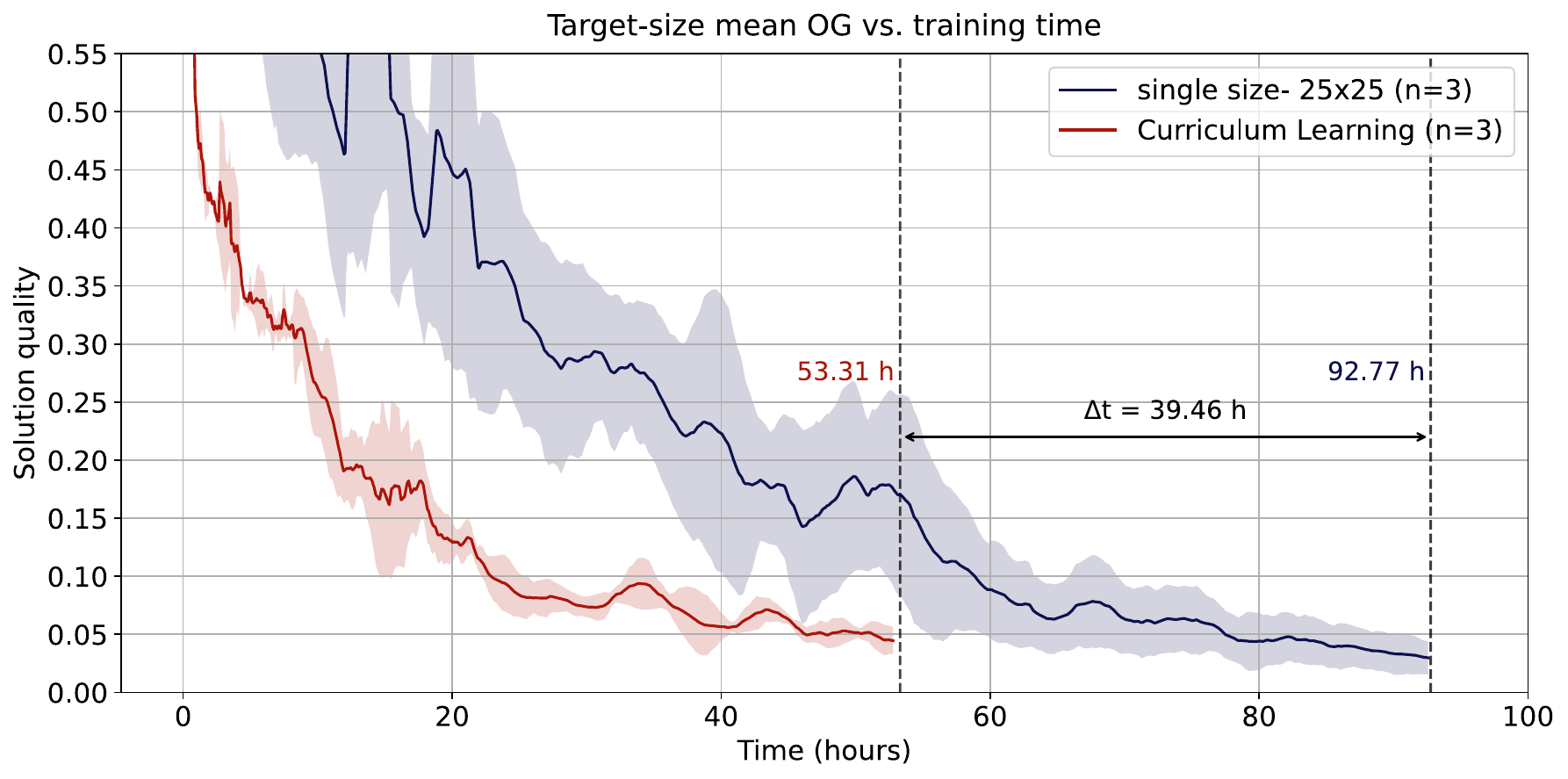}

    \caption{Specialization performance for the $25 \times 25$ target setting. Top: mean OG on $25 \times 25$ evaluation instances versus training steps. Bottom: mean OG on $25 \times 25$ evaluation instances versus wall-clock training time.}
    \label{fig:specialization_25}
\end{figure}

For the $25 \times 25$ setting, CL improves generalization while providing a substantial training-time reduction. As shown in Fig.~\ref{fig:generalization_25}, CL reduces the mean OG across all evaluation sizes by approximately 1.6 percentage points compared to 
single-size training, indicating improved performance over the mixed evaluation set. However, when the evaluation is restricted to the target size, CL shows a small specialization loss. As shown in Fig.~\ref{fig:specialization_25}, CL increases the 
target-size mean OG on $25 \times 25$ instances by approximately 1.3 percentage points compared to the single-size baseline. This suggests that, although CL improves cross-size generalization, direct training on $25 \times 25$ still provides a slight specialization 
advantage. In terms of efficiency, Fig.~\ref{fig:generalization_25} shows that CL reduces wall-clock training time by approximately 40 hours, corresponding to about a 42\% reduction compared to single-size training. Overall, the $25 \times 25$ results 
show that CL improves generalization and substantially reduces training time, while introducing a small specialization trade-off.

\subsection{Results for $30 \times 30$ Target Size}

\begin{figure}[t]
    \centering

    \includegraphics[
        width=\columnwidth,
        height=0.22\textheight,
        keepaspectratio
    ]{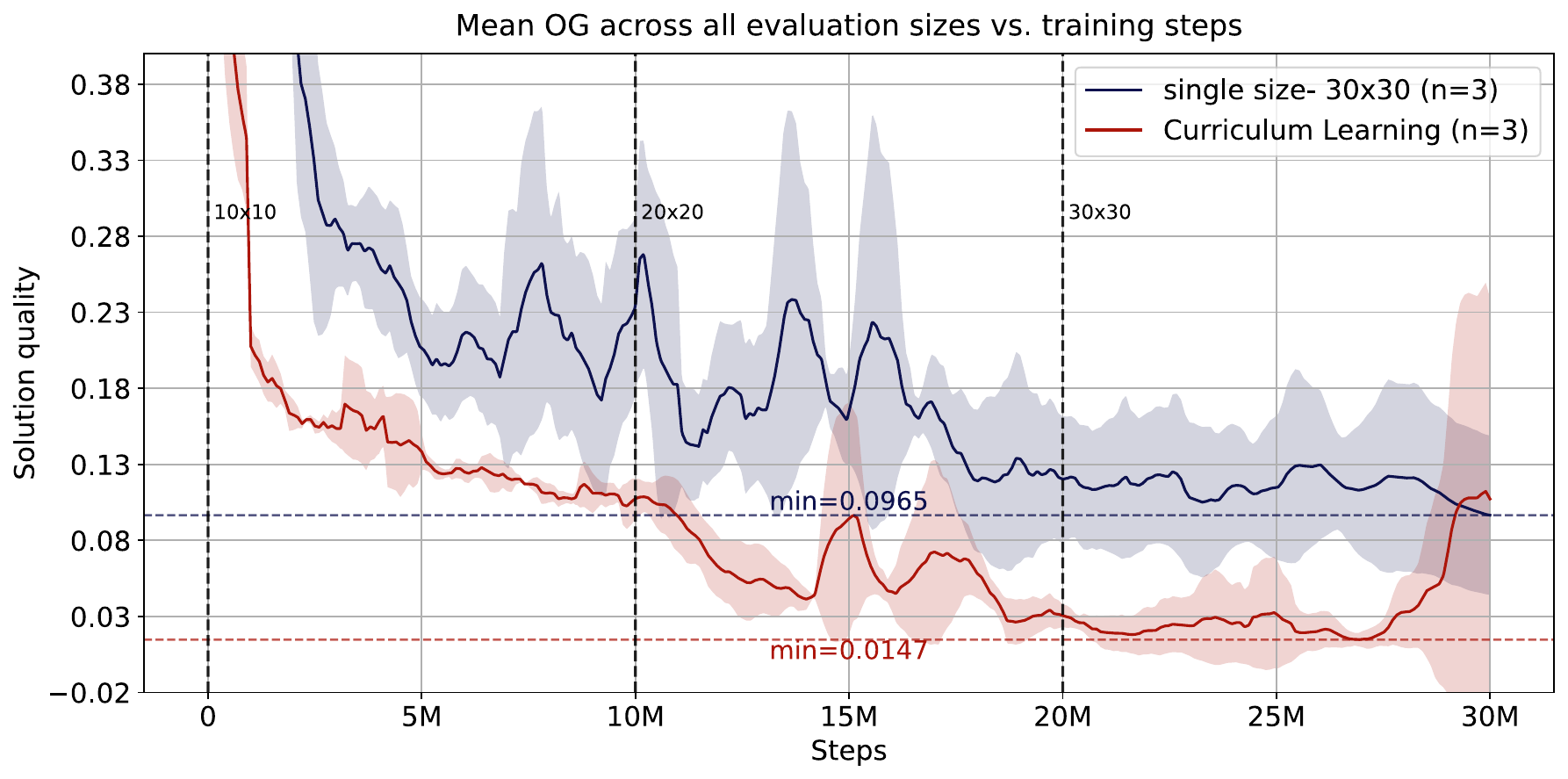}

    \vspace{1mm}

    \includegraphics[
        width=\columnwidth,
        height=0.22\textheight,
        keepaspectratio
    ]{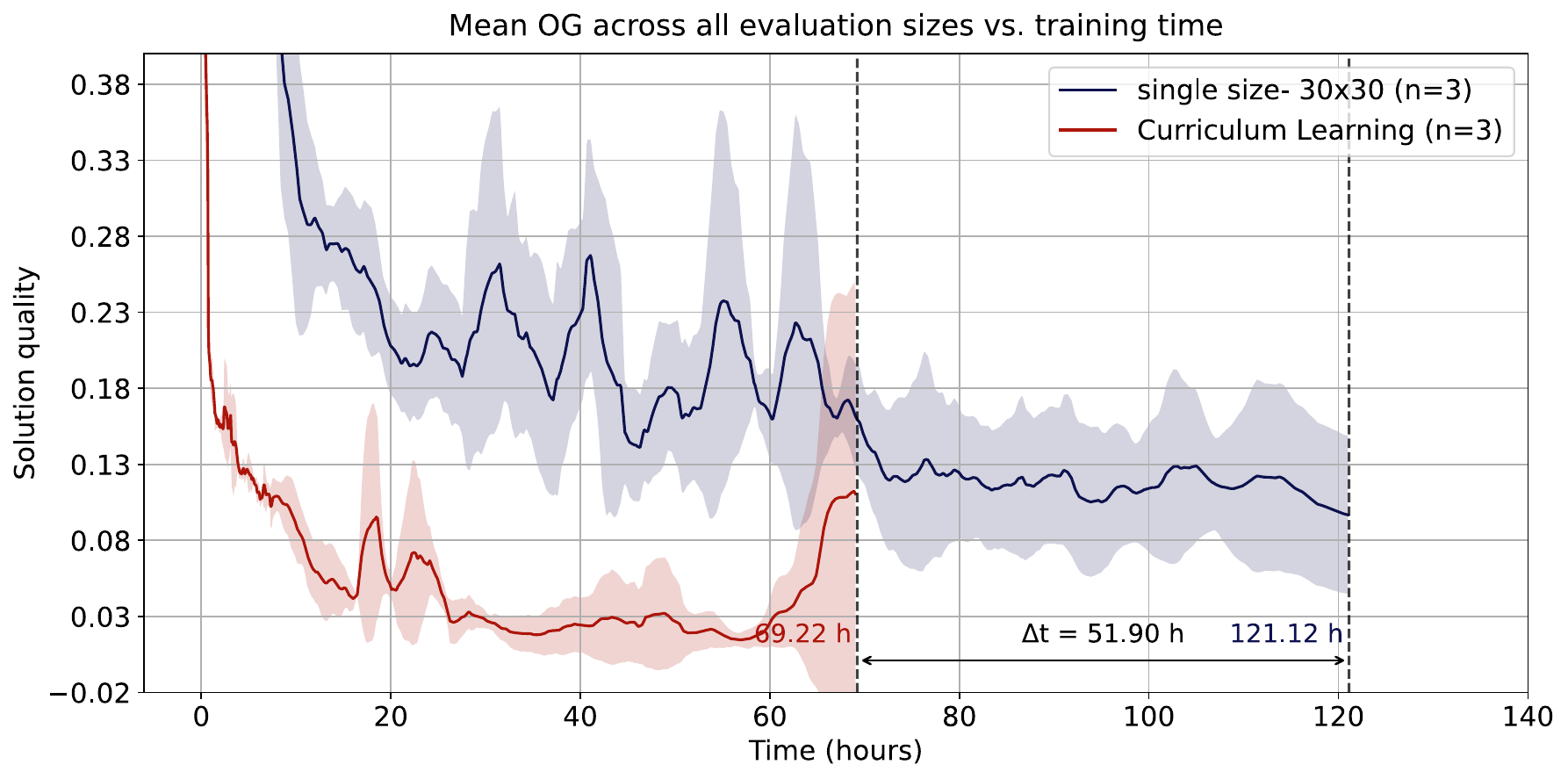}

    \caption{Generalization performance for the $30 \times 30$ target setting. Top: mean OG across all evaluation sizes versus training steps. Bottom: mean OG across all evaluation sizes versus wall-clock training time.}
    \label{fig:generalization_30}
\end{figure}

\begin{figure}[t]
    \centering

    \includegraphics[
        width=\columnwidth,
        height=0.22\textheight,
        keepaspectratio
    ]{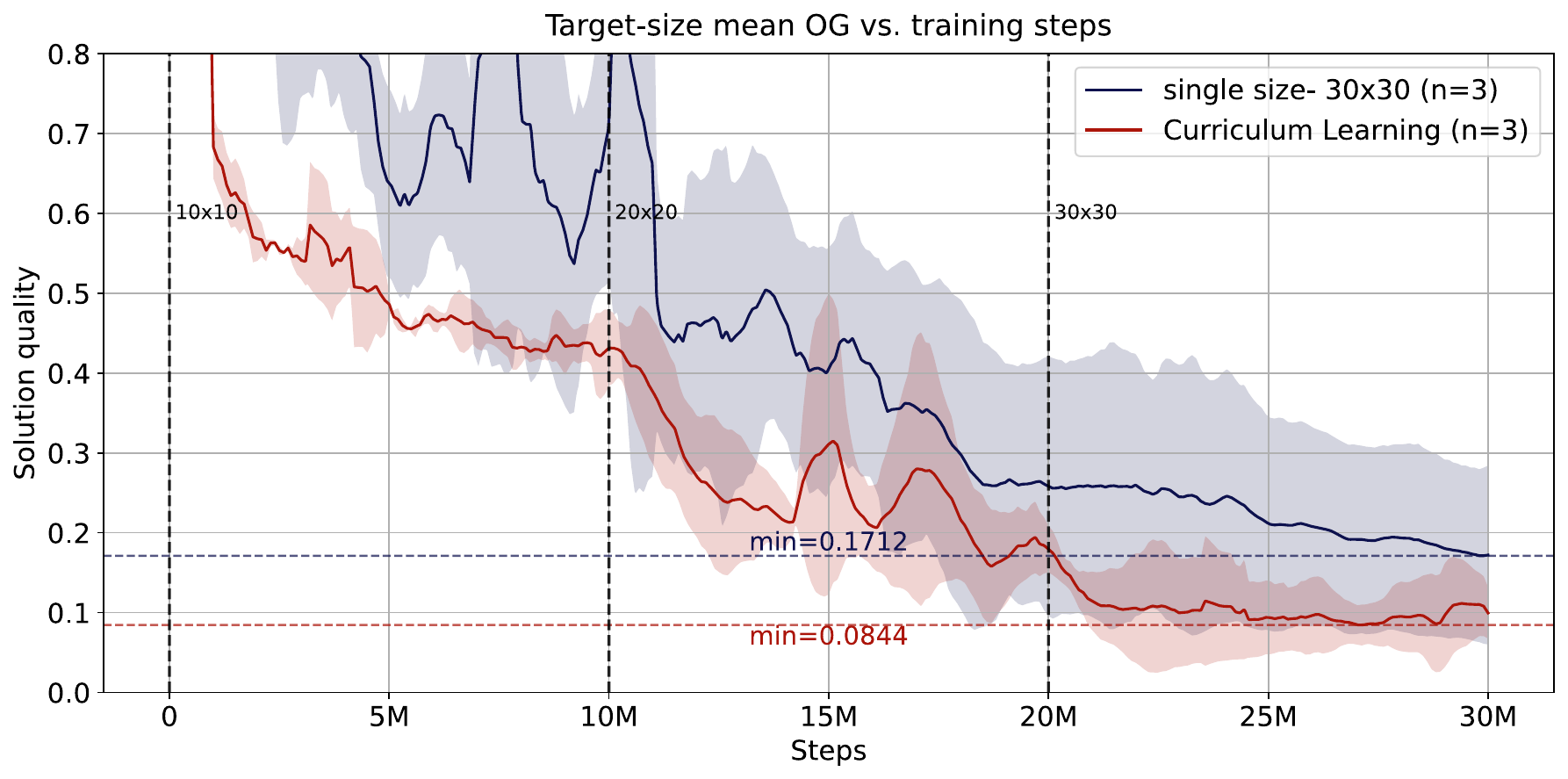}

    \vspace{1mm}

    \includegraphics[
        width=\columnwidth,
        height=0.22\textheight,
        keepaspectratio
    ]{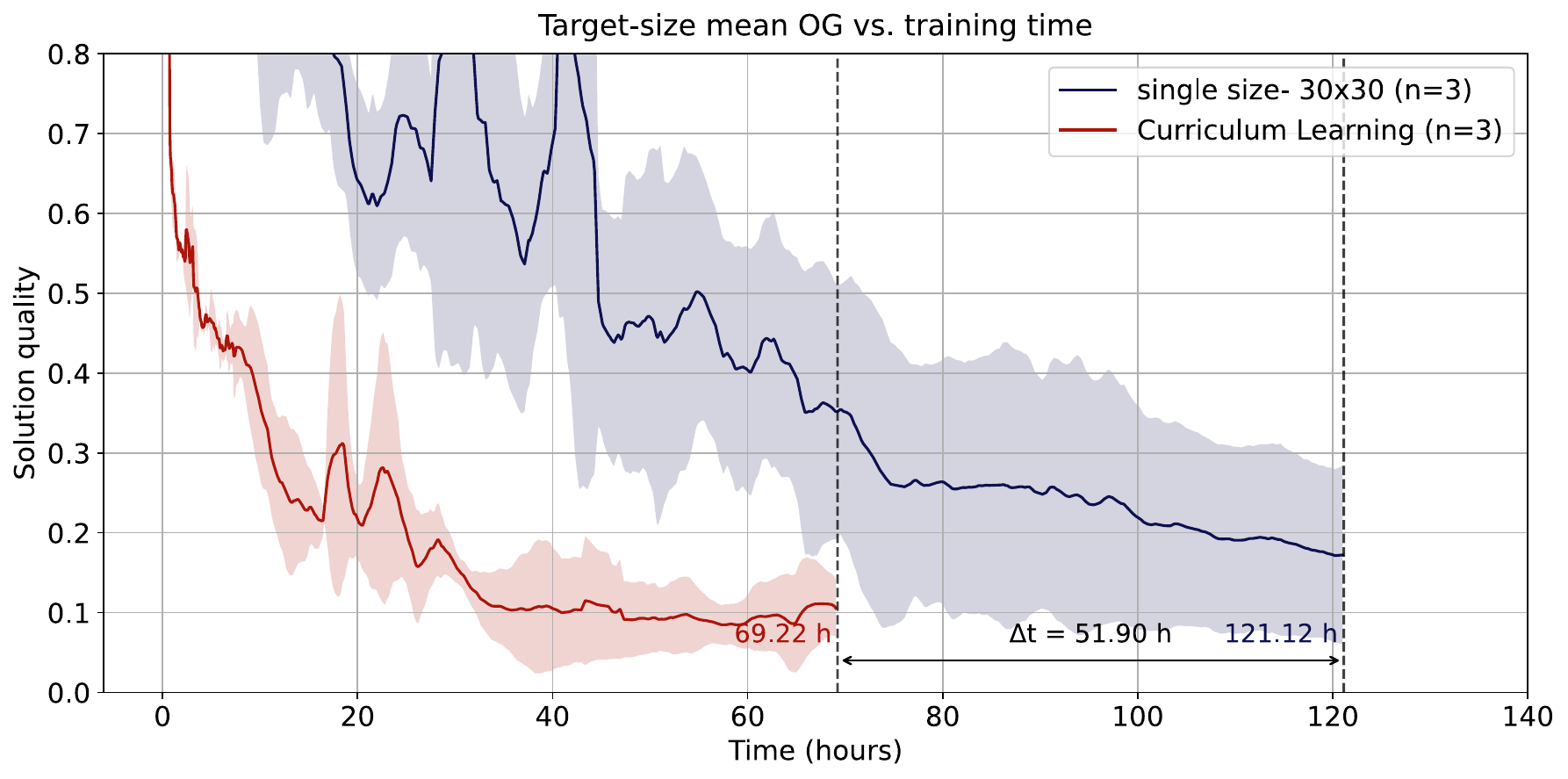}

    \caption{Specialization performance for the $30 \times 30$ target setting. Top: mean OG on $30 \times 30$ evaluation instances versus training steps. Bottom: mean OG on $30 \times 30$ evaluation instances versus wall-clock training time.}
    \label{fig:specialization_30}
\end{figure}

The strongest training-efficiency benefit is observed in the $30 \times 30$ setting, where direct training is most computationally expensive. As shown in Fig.~\ref{fig:generalization_30}, CL saves approximately 50 hours of wall-clock training time, 
corresponding to about a 43\% reduction compared to single-size training. This is the largest absolute time saving among the tested target sizes. In terms of solution quality, Fig.~\ref{fig:generalization_30} shows that CL reduces the mean OG across 
all evaluation sizes by approximately 8.1 percentage points. When the evaluation is restricted to the target size, CL also reduces the target-size mean OG by approximately 8.6 percentage points compared to the single-size baseline, as shown in 
Fig.~\ref{fig:specialization_30}. Thus, unlike the smaller target settings where CL shows a small specialization loss, the $30 \times 30$ setting improves performance on the target size.

Although the relative time reduction is only slightly higher than in the $25 \times 25$ setting, the absolute time saving is largest for $30 \times 30$. A possible explanation is that the $30 \times 30$ curriculum uses $20 \times 20$ as an intermediate 
stage, which is already more expensive than the $15 \times 15$ intermediate stage used in the $20 \times 20$ and $25 \times 25$ curricula. Overall, the $30 \times 30$ results provide the strongest evidence for CL, as it achieves the largest absolute 
time saving while improving both generalization and specialization, supporting the hypothesis that CL becomes more beneficial as the target instance size increases.

\begin{table}[t]
\centering
\caption{Summary of CL Compared to Single-Size Training}
\label{tab:summary_results}
\resizebox{\columnwidth}{!}{
\begin{tabular}{|c|c|c|c|}
\hline
\textbf{Target size} & \textbf{Generalization} & \textbf{Specialization} & \textbf{Time saved} \\
\hline
$20 \times 20$ & Similar & $\sim$1.2\% higher OG & $\sim$20 hours \\
\hline
$25 \times 25$ & $\sim$1.6\% lower OG & $\sim$1.3\% higher OG & $\sim$40 hours \\
\hline
$30 \times 30$ & $\sim$8.1\% lower OG & $\sim$8.6\% lower OG & $\sim$50 hours \\
\hline
\end{tabular}
}
\vspace{-3mm}
\end{table}

\section{Conclusion}
\label{sec:conclusion}

CL reduces training time for all target sizes, and the absolute time savings increase as the target size becomes larger. In terms of solution quality, CL mainly improves generalization performance, while specialization depends on the target size. 
For $20 \times 20$, CL achieves similar generalization but shows a small specialization loss, with the target-size mean OG increasing by approximately 1.2 percentage points. For $25 \times 25$, CL reduces the mean OG across all evaluation sizes by approximately 
1.6 percentage points, while the target-size mean OG increases by approximately 1.3 percentage points compared to single-size training. For $30 \times 30$, CL provides the strongest overall result, reducing the mean OG across all evaluation sizes by approximately 8.1 percentage 
points and the target-size mean OG by approximately 8.6 percentage points.

These observations suggest that CL is particularly useful when direct training on the target size becomes computationally expensive. Smaller instances provide shorter episodes and more frequent learning opportunities, allowing the agent to learn useful 
scheduling behavior before adapting to larger instances. When training progresses to more complex target sizes, the previously learned policy can be refined instead of being learned from scratch. The results also indicate that CL may introduce a small 
specialization trade-off for smaller and medium target sizes, since the single-size model spends the full training budget on the target size, whereas the curriculum model only trains on the target size during the final stage. However, as the target size 
increases, the generalization, specialization, and training-efficiency benefits become more pronounced. This suggests that CL over instance size can improve computational efficiency and cross-size generalization in GNN-based RL for JSSP, especially for 
larger target sizes. Table~\ref{tab:summary_results} summarizes the relative performance of CL compared to single-size training.

Several limitations remain. The curriculum switch interval is fixed at 10 M steps, and the study is limited to JSSP instances with $J/M = 1$. Future work should evaluate the approach on larger instances and different job-to-machine ratios. 
In addition, adaptive curriculum schedules based on convergence behavior, performance thresholds, or instance difficulty should be investigated. Moreover, while this study focuses on comparing size-based CL with direct single-size training, future work 
should include comparisons with adaptive curriculum strategies, difficulty-based curricula, and additional state-of-the-art scheduling methods to provide a broader evaluation.

\bibliographystyle{IEEEtran}
\bibliography{bibliography}

\end{document}